\documentclass[runningheads]{llncs}
\usepackage[T1]{fontenc}
\usepackage{graphicx}
\usepackage{booktabs}
\usepackage[misc]{ifsym}
\newcommand{\corr}{(\Letter)}
\usepackage{dsfont}
\usepackage{bm}

\usepackage{tikz}
\usetikzlibrary{scopes}
\usetikzlibrary{backgrounds}
\usetikzlibrary{shapes.geometric, patterns}
\pgfdeclarelayer{nodelayer}
\pgfdeclarelayer{edgelayer}
\pgfsetlayers{edgelayer,nodelayer,main}
\tikzstyle{new style 0}=[fill=white, draw=black, shape=rectangle]
\tikzstyle{none}=[]
\tikzstyle{Large Box}=[fill=none, draw=black, shape=rectangle, minimum width=7cm, minimum height=7cm]
\usepackage{soul}% for underlines
\tikzstyle{Edge arrow}=[->]

\usepackage{algorithm}
\usepackage{subcaption}
\usepackage{newfloat}
\usepackage{listings}

\usepackage{multirow}%
\usepackage{amsmath,amssymb,amsfonts}%
\usepackage{mathrsfs}%
\usepackage[title]{appendix}%
\usepackage{xcolor}%
\usepackage{textcomp}%
\usepackage{manyfoot}%
\usepackage{booktabs}%
\usepackage{algorithm}%
\usepackage{algorithmicx}%
\usepackage{algpseudocode}%
\usepackage{listings}%

\usepackage{siunitx}
\usepackage{booktabs}
\usepackage{collcell}
\usepackage{hyperref}

\newcommand{\quotes}[1]{``#1''}
\begin{document}

\title{Classifier-free graph diffusion for molecular property targeting}

\titlerunning{Classifier-free graph diffusion for molecular property targeting}

\author{Matteo Ninniri\inst{1} \corr  \and
Marco Podda\inst{1} \and
Davide Bacciu\inst{1}}

\authorrunning{M. Ninniri et al.}

\toctitle{Classifier-free graph diffusion for molecular property targeting}
\tocauthor{Matteo Ninniri,Marco Podda,Davide Bacciu}

\institute{Department of Computer Science - University of Pisa\\ Largo Bruno Pontecorvo, 3 - 56127 Pisa (Italy)\\ \email{m.ninniri1@studenti.unipi.it}\\
\email{\{marco.podda,davide.bacciu\}@unipi.it}}

\maketitle

\makeatletter
\def\blfootnote{\xdef\@thefnmark{}\@footnotetext}
\makeatother

\begin{abstract}
This work focuses on the task of property targeting: that is, generating molecules conditioned on target chemical properties to expedite candidate screening for novel drug and materials development. DiGress is a recent diffusion model for molecular graphs whose distinctive feature is allowing property targeting through classifier-based (CB) guidance. While CB guidance may work to generate molecular-like graphs, we hint at the fact that its assumptions apply poorly to the chemical domain. Based on this insight we propose a classifier-free DiGress (\textsc{FreeGress}), which works by directly injecting the conditioning information into the training process. CF guidance is convenient given its less stringent assumptions and since it does not require to train an auxiliary property regressor, thus halving the number of trainable parameters in the model. We empirically show that our model yields significant improvement in Mean Absolute Error with respect to DiGress on property targeting tasks on QM9 and ZINC-250k benchmarks. As an additional contribution, we propose a simple yet powerful approach to improve the chemical validity of generated samples, based on the observation that certain chemical properties such as molecular weight correlate with the number of atoms in molecules.

\end{abstract}

\section{Introduction}
%A short introduction to the subject
\blfootnote{This preprint has not undergone peer review (when applicable) or any post-submission improvements or corrections. The Version of Record of this contribution is published "in Machine Learning and Knowledge Discovery in Databases. Research Track", and is available online at https://doi.org/10.1007/978-3-031-70359-1\_19}

Generating molecules with desired chemical properties is crucial to enable fast candidate screening in the early stages of drug~\cite{Dara2021} or novel materials~\cite{Liu2017} development. Although the advent of deep learning has achieved remarkable accomplishments in computational chemistry, this strategic endeavor remains an area of active research. In the past, much of the researchers' focus has been on the \textit{property optimization} task, which is based on modifying a generated molecule such that it acquires some solicited chemical property~\cite{jbj:1}. However, driven by the success of conditional generative models~\cite{radford2021}, an equally important -- though slightly different -- task has emerged recently: \textit{property targeting}, that is, to natively generate molecules which satisfy pre-specified chemical desiderata~\cite{Sousa2021}.\\
\par Denoising Diffusion Probabilistic Models (DDPMs)~\cite{hja:1} have been shown to achieve state-of-the-art performance in conditional generation, for example to synthesize images from user-provided textual guidance~\cite{scslwdgamlshfn:1}. Due to their flexibility and excellent performance, DDPMs have been extended to the chemical domain, being applied to tasks such as distribution learning, property optimization, and property targeting with promising results~\cite{Runcie2023}.

DiGress~\cite{vkswcf:1} is one of the first successful applications of DDPMs to molecular generation. Under the hood, DiGress is based on a discrete diffusion process~\cite{ajhtb:1} which gradually applies noise to a molecular graph according to a transition matrix, while denoising is performed by a graph transformer network~\cite{Dwivedi2020}. The most interesting feature of DiGress is the possibility to perform conditional generation for property targeting through classifier-based (CB) guidance~\cite{dn:1}. Loosely speaking, CB guidance requires to train a separate classifier to predict the conditional information from noisy samples, and to inject the resulting gradients back into the reverse process to bias the generative process. While successful to some extent, CB guidance has been shown to be inherently limited by \textit{i}) the necessity of training a separate property predictor, which defies the purpose of having a single conditional generative model in the first place, and \textit{ii}) the fact that the gradients of the property predictor are not always informative and may lead the generative process astray. Due to these limitations, classifier-free (CF) guidance for DDPMs is often preferred. The idea behind CF guidance is to directly incorporate the conditioning vector as input to train the conditional DDPM. With respect to CB guidance, CF guidance has been shown to enable more stable training and better generative performance in general~\cite{hs:1}.

With DiGress for molecular generation specifically, CB guidance is further limited by the fact that the auxiliary model is a regressor whose predictions (the chemical properties) are assumed to be normally distributed even for noisy graphs which are chemically invalid.
This motivates our first contribution, which consists of the development and implementation of CF guidance for DiGress, called \textsc{FreeGress}. 
Experimentally, we show that switching from CB to CF guidance is beneficial to improve at property targeting. In particular, we evaluated \textsc{FreeGress} against DiGress on the QM9~\cite{Ramakrishnan2014} and ZINC-250k~\cite{GomezBombarelli2018} datasets, where we queried the models to generate molecules with properties as close as possible to a target specification. Comparing the mean absolute error between the target properties and the properties of the generated molecules, \textsc{FreeGress} significantly outperformed DiGress, with improvements up to $88\%$ in the most favourable case. Besides improving performance, \textsc{FreeGress} does not require an auxiliary property regressor, reducing the number of trainable parameters. Furthermore, guided by the observation that certain chemical properties relate to the molecular graph size (the number of atoms in the molecule), we also propose to improve the generative process by first learning the probability of sampling a certain number of nodes given the target property, and then using samples from this distribution to constrain the size of the graph to be generated. Through experiments, we show that this simple method allows to generate more chemically valid graphs without sacrificing performance. Our code and supplementary material are available at \url{https://github.com/Asduffo/FreeGress}.

\section{Background and related works}
\subsubsection{Notation.} 
We represent a molecular graph with $n$ atoms as a tuple $G=(\bm{X}, \bm{\textsf{E}})$ where $\bm{X} \in \mathbb{R}^{n \times a}$ is a node matrix whose $i$-th row, indicated as $\bm{x}_i \in \mathbb{R}^a$, is the one-hot encoding of the atomic type of the $i$-th atom. Possible atomic types are specified by the set $\mathcal{X}$ with cardinality $a$. Similarly, $\bm{\textsf{E}} \in \mathbb{R}^{n \times n \times b}$ is an adjacency tensor (also called edge tensor) that jointly represents the molecular connectivity and the type of bond occurring between each pair of atoms. The $(i,j)$-th entry of $\bm{\textsf{E}}$, indicated as $\bm{e}_{ij} \in \mathbb{R}^b$, contains the one-hot encoding of the bond type occurring between the $i$-th and $j$-th atoms. Possible bond types are specified by the set $\mathcal{E}$ with cardinality $b$, where the absence of a bond is itself considered a bond type. In the following we will use the term \quotes{node} interchangeably with \quotes{atom}, and \quotes{edge} interchangeably with \quotes{bond}. Conditioning vectors will be generally represented as vectors $\bm{y} \in \mathbb{R}^d$, where $d$ is usually a small integer.

\subsection{Deep generative models for molecules}
The first approach to use a deep generative model to produce unseen molecules has been proposed by~\cite{GomezBombarelli2018}. Essentially, the model is a VAE where both the encoder and decoder are recurrent neural networks trained to reconstruct SMILES strings.
Other methods related to this generative flavour use the alternative language of SELFIES instead, which is deemed to be more robust than SMILES as every possible character sequence defines a valid molecule. 
A different branch of methods is focused on generating the molecular graph directly. 
On graph-based models, property optimization has been performed with simple hill climbing in latent space (guided by a property predictor), up to iterative reinforcement learning approaches which assign higher rewards to chemically appealing molecules~\cite{you2018gpcn}.
Lately, DDPMs that generate the molecular graph have started to be used for distribution learning as well as property optimization tasks.

\subsection{Denoising Diffusion Probabilistic Models}
DDPMs consist of an untrained \textit{forward process} $q(\bm{x}^t|\bm{x}^{t-1})$ and a parameterized \textit{reverse process} $p_\theta(\bm{x}^{t-1}|\bm{x}^t)$. The former iteratively perturbs the initial data point $\bm{x}^0$ to transform it into an $\bm{x}^T$ akin to Gaussian noise, while the second is trained to incrementally remove the noise from $\bm{x}^T$ until $\bm{x}^0$ is restored. Generation from a trained DDPM amounts to sample $\bm{\tilde{x}}^T \thicksim \mathcal{N}(0, \bm{I})$ from an isotropic Gaussian $\mathcal{N}$ and iteratively applying the denoising model for $T$ steps until a new sample $\bm{\tilde{x}}^0 \thicksim q$ is obtained.
Conditioned generation in DDPMs is achieved by injecting a \textit{guidance vector}, or \textit{guide} $\bm{y}$, to obtain a \textit{conditioned reverse process} $p_{\theta}(\bm{x}^{t-1}|\bm{x}^t, \bm{y})$. DDPMs defined as above are not suited for graphs, since Gaussian noising does not work effectively on discrete distributions. 

\subsubsection{Classifier-based guidance} CB guidance~\cite{dn:1} refactors the reverse process as $p_{\theta}(\bm{x}^{t-1}|\bm{x}^t)p_{\phi}(\bm{y}|\bm{x}^{t-1})$, where $p_{\phi}(\bm{y}|\bm{x}^{t-1})$ employs an auxiliary classifier trained to predict the guide from a noisy version of the input. A scaled version of the regressor gradient $\nabla_{\bm{x}^t}$ is used to magnify the conditioning signal. CB guidance is limited by the fact that it requires to train the auxiliary classifier, without the possibility of exploiting pre-trained models; moreover, since only few parts of the noisy $\bm{x}^t$ are actually useful to predict $\bm{y}$, the resulting gradient could yield undesirable directions in input space. 

\subsubsection{Classifier-free guidance}
Differently from the CB approach, CF guidance~\cite{hs:1} jointly optimizes $p_{\theta}(\bm{x}^{t-1}|\bm{x}^t, \bm{y})$, the conditioned model, and $p_{\theta}(\bm{x}^{t-1}|\bm{x}^t)$, the unconditioned model, at the same time.  During sampling, the reverse process is computed as the barycentric combination $(1 + w) \, p_{\theta}(\bm{x}^{t-1}|\bm{x}^t, \bm{y}) - w \, p_{\theta}(\bm{x}^{t-1}|\bm{x}^t)$ between the conditioned and unconditioned predictions with weight $w \in \mathbb{R}_+$.

\subsection{DiGress: Denoising Diffusion for Graphs}
Here we briefly describe DiGress~\cite{vkswcf:1}, a discrete DDPM for graphs which is the starting point of this study. 
In the forward process, DiGress injects noise in the data at time-step $t-1$ by multiplying it with a transition matrix, which loosely speaking specifies the probability of transitioning from one node type to another. The forward process is applied until the training graph is completely corrupted.
The reverse process implemented by DiGress is specified as $p_{\theta}(G^{t-1}|G^t)$, which is factored (by assuming node and edge independence) as a product of conditionals of the individual nodes and edges given the current graph. In turn, each node conditional is obtained by marginalizing over the node types (resp. edge types for the edge conditionals). Specifically, the nodes (resp. edge) marginal is computed by predicting the true node (resp. edge) types from their noisy intermediates.
For conditioned generation, DiGress employs CB guidance. Specifically, the conditioned reverse process is formulated as:
\begin{equation}
    p_{\theta}(G^{t-1}|G^t)p_{\phi}(\bm{y}|G^{t-1}),
\end{equation}
where $p_{\phi}(\bm{y}|G^{t-1}) \propto \exp(-\lambda \langle \triangledown_{G^t} ||\bm{y} - g_{\phi}(G^t) ||^2, G^{t-1} \rangle)$, $g_{\phi}$ is a property regressor, and $\lambda$ is a hyper-parameter that magnifies the conditional gradient.

\subsubsection{Limitations} While the development of unconditioned DiGress appears intuitive, the use of CB guidance for the conditional variant leads to design choices and assumptions that are unrealistic in the chemical domain. Firstly, CB guidance relies on an external predictor that learns chemical properties from noisy graphs, which implies that chemically invalid molecules are unnaturally related to valid ones by being assigned the same properties. Secondly, the distribution learned by the auxiliary property regressor is assumed to be Normal with mean $p(\bm{y}|G^t)$, approximated by $g_{\phi}$ by minimizing a mean squared error objective. This assumption is unsupported by premise for most denoising steps (where $G^t$ usually represents an invalid molecule), since it implies a distribution of chemical properties for an object that does not even exist in chemical space. Moreover, even if two molecules with graphs $G^{t}$ and $G^{t-1}$ are both valid and similar, their chemical properties can be drastically different, since molecular property landscapes are in general non-smooth~\cite{Aldeghi2022}, violating again the normality assumption. This discussion motivates our intention to develop CF guidance for DiGress with the objective of making the conditional process simpler and at the same time appropriate for the chemical context where it is applied.
\begin{figure}[!htb]
\makebox[\textwidth][c]{
	\begin{tikzpicture}[node distance={30mm}, thick, main/.style = {draw, circle}] 
		\node[] (1)              {\includegraphics[width=2cm, trim={0 2cm 0cm 0cm},clip]{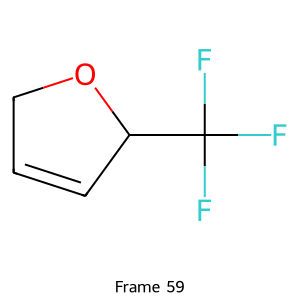}}; 
		\node[] (2) [right of=1] {\includegraphics[width=2cm, trim={0 2cm 0cm 0cm},clip]{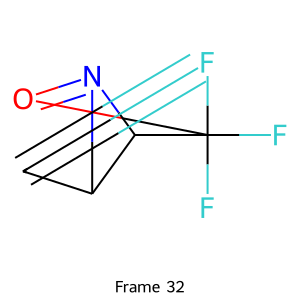}}; 
		\node[minimum size=1.75cm] (3) [right of=2] {...}; 
		\node[] (4) [right of=3] {\includegraphics[width=2cm, trim={0 2cm 0cm 0cm},clip]{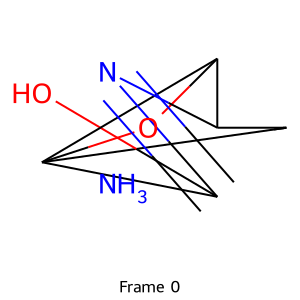}}; 
        \node[] (5) [node distance={7.5mm}, below of=4] {$n \sim p_{\varepsilon}(n \mid \bm{y}); \bm{y}$};
        \node[] (6) [node distance={7.5mm}, above of=1] {$G^0 = (\bm{X}^0, \bm{\textsf{E}}^0)$};
		
		\draw[->] (1) to [out=80,in=100,looseness=0.7] node[above, align = center]{$\bm{X}^1 \sim Cat(\bm{X}^0Q^1_{\bm{X}})$ \\ $\bm{\textsf{E}}^1 \sim Cat(\bm{E}^0Q^1_{\bm{\textsf{E}}})$} (2);
		\draw[->] (2) to [out=80,in=100,looseness=0.7] node[above]{ } (3);
		\draw[->] (3) to [out=80,in=100,looseness=0.7] node[above, align = center]{$\bm{X}^T \sim Cat(\bm{X}^{T-1}Q^T_{\bm{X}})$ \\ $\bm{\textsf{E}}^T \sim Cat(\bm{\textsf{E}}^{T-1}Q^T_{\bm{\textsf{E}}})$} (4);
		
		\draw[->] (4) to [out=260,in=280,looseness=0.7] node[below]{$(\bm{X}^{T-1}, \bm{\textsf{E}}^{T-1}) \sim p_\theta(G^{T-1}|G^T, \bm{y})$} (3);
		\draw[->] (3) to [out=260,in=280,looseness=0.7] node[below]{ } (2);
		\draw[->] (2) to [out=260,in=280,looseness=0.7] node[below]{$(\bm{X}^{0}, \bm{\textsf{E}}^{0}) \sim p_\theta(G^0|G^1, \bm{y})$} (1);
	\end{tikzpicture}
}

\caption{A depiction of \textsc{FreeGress}. The forward process, which gradually corrupts a molecule into a random graph, goes from left to the right. The reverse process, which denoises the original graph, goes from right to left. Note that the reverse process allows for a conditioning vector $\bm{y}$ and a number of nodes $n$ sampled from a trained neural network $p_{\varepsilon}$.}
\label{fig:diffusion_schema}
\end{figure}

\section{Classifier-Free Graph Diffusion}
This section presents our first contribution, a classifier-free DDPM for conditioned molecular generation named \textsc{FreeGress}. A high-level summarization of the proposed model is sketched in Figure \ref{fig:diffusion_schema}.

\subsubsection{Forward process} The forward process of \textsc{FreeGress} is analogous to that of DiGress, and essentially consists of applying $T$ noising steps to an input molecular graph $G$ according to the following process:
\begin{equation} \label{eq:g^t_given_g^t_min^1}
	q(G^t|G^{t-1}) = (\bm{X}^{t-1}\bm{Q}^t_{\bm{X}}, \bm{\textsf{E}}^{t-1}\bm{Q}^t_{\bm{\textsf{E}}}), \; 1 \leq t \leq T
\end{equation}
where $\bm{Q}^t_X$ and $\bm{Q}^t_{\bm{\textsf{E}}}$ are transition matrices applied to the node matrix and the edge tensor, respectively. In particular, $\bm{Q}^t_X$  is formulated as:
\begin{equation} \label{eq:transition_matrix}
    \bm{Q}^t_{\bm{X}} = \alpha^t\bm{I} + \beta^t \bm{1}_{a}\bm{m}_{\bm{X}},
\end{equation}
where $\bm{m}_{\bm{X}}$ is a vector containing the marginal distribution of the node types in the training set, $\alpha^t$ is a noise scheduler, and $\beta^t = 1 - \alpha^t$. In practice, the transition probability to a certain node type is given by its frequency in the training set. The transition tensor $\bm{Q}^t_{\bm{\textsf{E}}}$ is chosen analogously.  Unrolling the forward process equation, we can sample $G^t$ given $G^0$ in a single step as follows:
\begin{align}
q(G^t|G^0) = (\bm{X}\overline{\bm{Q}}^t_{\bm{X}}, \bm{\textsf{E}}\overline{\bm{Q}}^t_{\bm{\textsf{E}}}),
\end{align}
where  $\overline{\bm{Q}}^t_{\bm{X}} =   \overline{\alpha}^t\bm{I} + \overline{\beta}^t \bm{1}_{a}\bm{m}_{\bm{X}}$, $\overline{\alpha}^t = \prod_{\tau = 1}^{t} \alpha^\tau$, and $\overline{\beta}^t = 1-\overline{\alpha}^t$. Notice that  $\overline{\bm{Q}}^t_{\bm{\textsf{E}}}$ is defined analogously.

\subsubsection{Reverse Process}
The key impact of CF guidance is instead on the adaptive reverse process. To this end, we begin by rewriting the conditioned reverse process as:
\begin{equation}
    p_\theta(G^{t-1}|G^t, \bm{y}),
\end{equation}
such that it now incorporates the guide $\bm{y}$ as input. Then, we assume independence between nodes and edges and factorize as the following product of conditionals:
\begin{align}
	 p_\theta(G^{t-1}|G^t, \bm{y}) = \prod_{1 \leq i \leq n} p_\theta(\bm{x}_i^{t-1}|G^t, \bm{y}) \prod_{1 \leq i,j \leq n} p_{\theta}(\bm{e}_{ij}^{t-1}|G^t, \bm{y}).
\end{align}
Following~\cite{gcbwzcyg:1}, we rewrite the nodes marginal as:

\begin{equation} \label{eq:xtm1_formula}
p_\theta(\bm{x}_i^{t-1}|G^t, \bm{y}) = \sum_{\bm{x} \in \mathcal{X}} q(\bm{x}_i^{t-1}|\bm{x}_i^{t}, \bm{x}_i^0 = \bm{x})\,f_{\theta}(\bm{x}^{0}_i=\bm{x}|G^t, \bm{y}),
\end{equation}
where $f_{\theta}$ is a neural network that predicts the true node types $\bm{x}^0$ using both the current noisy graph $G^t$ and the guide $\bm{y}$. Similarly, the edges marginal is:
\begin{align}
	 p_{\theta}(\bm{e}_{ij}^{t-1}|G^t, \bm{y}) = \sum_{\bm{e} \in \mathcal{E}} q(\bm{e}_{ij}^{t-1}|\bm{e}_{ij}^{t}, \bm{e}_{ij}^0 = \bm{e})\,f_{\theta}(\bm{e}_{ij}^{0}=\bm{e}|G^t, \bm{y}).
\end{align}

At this point, we have, in principle, a classifier-free DDPM for graphs. However, conditioning might not work as expected in this scenario, since the model could learn to completely ignore the guide $\bm{y}$. This behaviour has been observed, e.g., by~\cite{tgbcw:1}. To prevent this, we leverage \textit{conditional dropout}~\cite{hs:1}, where the conditioning vector $\bm{y}$ is replaced, with probability $\rho$, by a non-parametrized placeholder $\bm{\bar{y}}$. 
To summarize, the final vector $\bm{y}$ employed in the equations (and, consequently, in the neural network) is chosen as follows:
\begin{equation}
		\bm{y} \leftarrow \begin{cases}
               \bm{\bar{y}} & \text{with probability } \rho\\
               \bm{y}       & \text{with probability } 1 - \rho
		\end{cases}
\end{equation}
Intuitively, the placeholder signals the network to perform inference without a guidance vector, as if the process was unconditioned. As a result, the model becomes capable of performing both conditioned and unconditioned sampling. To complete the derivation of \textsc{FreeGress}, we build on~\cite{hs:1} and express $f_{\theta}$ as:

\begin{equation} \label{eq:inference}
	p_{\theta}(\bm{x}^{0}_i|G^t) + s(p_{\theta}(\bm{x}^{0}_i|G^t,\bm{y}) - p_{\theta}(\bm{x}^{0}_i|G^t))
\end{equation}
which is computable since the model can output the unconditioned probability using the placeholder in place of the guide or, more formally, since ${p(\bm{x}^{0}_i|G^t)} \approx {p_{\theta}(\bm{x}^{0}_i|G^t, \bm{\bar{y}}})$. The process is symmetric for the edges $\bm{e} \in \mathcal{E}$. Finally, to add more flexibility, we follow~\cite{tgbcw:1} and we parametrize $\bm{\bar{y}}$ so that it is trainable.

The major difference between~\cite{hs:1} and our work is the fact that the former is designed for continuous signals, such as images, and their neural network is trained to predict the Gaussian noise $\epsilon$ corrupting the input signal $\bm{x}^t$ given both the latter and the conditioning vector $\bm{y}$. \textsc{FreeGress}, instead, is designed for discrete graphs and its neural network is trained to infer $G^0$ given the corrupted graph $G^t$ and conditioning vector $\bm{y}$. An alternative formula, proposed by~\cite{tgbcw:1}, is shown in Equation~\ref{eq:log_inference}:
\begin{equation} \label{eq:log_inference}
    \log{p(\bm{x}^0_i|G^t)} + s(\log{p(\bm{x}^0_i| G^t,\bm{y})} - \log{p(\bm{x}^0_i|G^t)}).
\end{equation}
Note how setting the hyper-parameter $s$ to zero is equal to disabling the conditional dropout altogether.
\input{figures/model}
\subsubsection{Neural Architecture.}

Figure~\ref{fig:architecture} depicts the neural architecture of our approach integrating explicit guidance. The network consists of a stack of Graph Transformer layers. The input matrices $\bm{X}$ and $\bm{\textsf{E}}$ can be augmented with arbitrary node and edge features deemed useful (indicated with the $+$ superscript in Figure~\ref{fig:architecture}). In addition to the two matrices, a further vector $\bm{u}$ is used as input, containing the current timestep $t$ along with optional graph-level features (such as the overall molecular weight).
All these data are augmented with the guidance $\bm{y}$ and processed by the stack of graph transformer layers. The graph transformer is displayed in Figure~\ref{fig:graph-transformer}: at a high level, it resembles a standard transformer architecture, with a self attention module followed by dropout, residual connection, layer norm, and a feedforward stack. The self attention layer (shown in Figure~\ref{fig:self-attention}) first processes the (augmented) $\bm{X}$ matrix through the standard outer product and scaling typical of attention layers. Then, the result is used in conjunction with the edge tensor $\bm{\textsf{E}}$  as input to a FiLM layer~\cite{psvdc:1} to produce unnormalized attention scores. The scores are passed through a softmax layer, combined once again with the values $\bm{X}$. The attention vectors are then flattened and mixed with the vector $\bm{u}$ through another FiLM layer, and then transformed linearly to obtain the prediction logits. The vector $\bm{u}$ is instead processed differently, as shown visually in the rightmost part of Figure~\ref{fig:self-attention}. Specifically, $\bm{X}$ and $\bm{\textsf{E}}$ are first passed through independent PNA layers~\cite{corso2020principal} and then summed together with a linear projection of $\bm{u}$ to form the final representation. 

\subsubsection{Loss function.} \label{sec:loss}
During training, \textsc{FreeGress} optimizes the following loss function for a given time step $t$:
\begin{equation}
\sum_{1 \leq i \leq n} \mathrm{CE}\left(\bm{x}_i^0, f_{\theta}(\bm{x}_i^t)\right)+ \gamma \sum_{1 \leq i, j \leq n} \mathrm{CE}\left(\bm{e}_{i j}^0, f_{\theta}(\bm{e}_{ij}^t)\right),
\end{equation}
which is the cross-entropy (CE) between the actual node (resp. edge) types and those predicted by the neural network described above. $\gamma$ is a hyper-parameter which adjusts the preference towards edges prediction. When Equation~\ref{eq:log_inference} is employed, CE is replaced by the negative log-likelihood loss.

\subsubsection{Conditioned Inference.} At each reverse process iteration, \textsc{FreeGress} generates a slightly denoised graph by first computing from the network $p_{\theta}(\bm{x}_i^{0}|\bm{x}_i^{t}, \bm{y})$, the conditioned probability, and $p_{\theta}(\bm{x}_i^{0}|\bm{x}_i^{t}, \bm{\bar{y}})$, the unconditioned one. The final probability is obtained by summing the two terms while giving an arbitrary (usually larger) weight $s$ to the conditioned probability, using Equation~\ref{eq:inference}. Then, the posteriors $q(\bm{x}_i^{t-1}|\bm{x}_i^{t}, \bm{x}_i^0 = \bm{x})$ are computed $\forall \bm{x} \in \mathcal{X}$ and summed as in Equation~\ref{eq:xtm1_formula}. This procedure is executed for each atom in the graph, and the resulting probabilities are sampled to obtain the nodes of the graph $G^{t-1}$. The process is symmetric for the graph edges.

\subsection{Conditioning on the number of nodes}\label{sec:node-inference}
A general limitation of diffusion models is the fact that the size of a sample (for graphs specifically, the number of nodes) cannot change during the denoising process. Some studies consider the possibility of inserting and removing elements from a generated, mono-dimensional sequence~\cite{jabt:1} but, to the best of our knowledge, there are no similar works for graphs. In the original implementation of DiGress, the number of nodes the generated graph will have is sampled from the marginal distribution computed from the training and validation sets, respectively. While this is not an issue for unconditioned sampling, we observed that for conditioned sampling, a certain property (e.g. molecular weight) might be featured only by molecules with a specific number of atoms. To solve this problem, we propose that the number of nodes $n$ of the graph to be generated is sampled from $p_{\varepsilon}(n|\bm{y})$, which is parametrized as a neural network with two hidden layers and a softmax output layer. Here, the idea is to exploit the guide even before generation starts, by providing a graph size which is correlated with the requested property.

\section{Experiments}
Here, we detail the experimental analysis to evaluate \textsc{FreeGress} on property targeting tasks by generating compounds that meet pre-specified properties.

\subsection{Datasets} 
Our experiments were conducted on QM9, a dataset of 133k small molecules (up to 9 heavy atoms) which was also part of the evaluation of DiGress, and ZINC-250k, a collection of 250k drug-like molecules selected from the ZINC dataset. For the latter, the molecules were first preprocessed by removing stereochemistry information. We have also removed scarcely present non-neutrally-charged atoms, leaving only N+ and O-. The latter have been treated as standalone atom types instead of their neutrally-charged counterparts. The final dataset size was consequently reduced to 228k molecules. Further statistics about the datasets are provided in the Supplementary Section 1.

\subsection{Metrics and targets}\label{sec:metrics}
All the experiments were performed using the following setup. After we trained each model, we randomly sampled 100 molecules from the dataset and computed the desired target properties as vectors $\bm{y}$, which were used as conditioning vectors. Then, we used the vectors 10 times each to perform conditioned generation, for a total of 1000 generated molecules for each model. We computed the desired properties on the generated samples\footnote{Using packages such as \texttt{RDKit}~\cite{l:1} and \texttt{psi4}~\cite{tsphefmbwarljswskvsc:1}} and compared the results with the properties of the original samples. The metric chosen for the comparison is the Mean Absolute Error (MAE) between the target properties and the properties of the generated molecules:
$$\frac{1}{1000}\sum_{i=1}^{100}\sum_{j=1}^{10}|\bm{y}_i - \widehat{\bm{y}}_{i,j}|,$$
where $\bm{y}_i$ are the target properties of the $i$-th molecule from the dataset and $\widehat{\bm{y}}_{i,j}$ is the target property of the $j$-th molecule generated using the properties of the $i$-th molecule as guide.
The targeted properties were the dipole moment $\mu$ and the Highest Occupied Molecular Orbital (HOMO) for QM9; for ZINC-250k, we targeted the Log-Partition coefficient (LogP), and the Quantitative Estimation of Drug-likeness (QED). The proposed node inference method was evaluated on the ZINC-250k dataset targeting Molecular Weight (MW).

\subsection{Experimental details}
\subsubsection{QM9 for $\mu$/HOMO targeting.} We trained and compared a set of \textsc{FreeGress} variants against a set of DiGress models, including an unconditional variant as baseline. To the best of our knowledge, DiGress is the only model that allows for property targeting at the time of writing.
\textsc{FreeGress} instances were trained with $s=\{1, 3, 5\}$, and with $\rho = \{0, 0.1, 0.2\}$ to study the effect of conditional dropout on the generative process. 
We have also experimented variants with additional features and $\rho = \{0, 0.1\}$; in this case, $\rho = 0.2$ was excluded as it did not improve on the former in preliminary trials. For DiGress, we evaluated the performance using $\lambda = \{200, 400, 600, 800\}$. To ensure a fair comparison, we trained both DiGress and \textsc{FreeGress} with matching architectural design: specifically, we used 5 graph transformer layers; embedding size of 256 for $\bm{X}$ and 128 for both $\bm{\textsf{E}}$ and $\bm{u}$; 8 attention heads; $T=500$, 1200 training epochs with batch size 512; Amsgrad optimizer~\cite{sashank} with learning rate of $2^{-4}$ and weight decay of $1^{-12}$; $\gamma = 2$. Importantly, since \textsc{FreeGress} does not require an auxiliary regressor, all the variants are trained with half the number of parameters than DiGress (4.6 million vs. 9.2 million, approximately). A single training required up to 12 hours on an nVidia V100 with 16GB of VRAM. 

\subsubsection{ZINC-250k for logP/QED targeting.} For these experiments we used a slightly different setup. Specifically, we trained from scratch both DiGress and \textsc{FreeGress} with 12 layers, 1000 training epochs, batch size 256, and doubling the embedding size of $\bm{y}$, while keeping the other hyper-parameters the same. We used larger models since ZINC-250k includes bigger (in terms of number of atoms) and more diverse molecules. In this case, we did not perform experiments with additional features since they increased  training time by a prohibitive amount. \textsc{FreeGress} variants required approximately 16 million parameters, while DiGress used approximately 32 million including the property regressor. Training the various models required approximately 5 days each on an nVidia A100 with 80GB of VRAM. 

\subsubsection{ZINC-250k for MW targeting.} These experiments evaluate the node inference method proposed in Section~\ref{sec:node-inference}. As such, we targeted MW since it trivially depends on the number of nodes. The setup is similar to the logP/QED experiments, with some slight differences. In this case only, we have managed to train \textsc{FreeGress} with additional features and $\rho \in \{0, 0.1\}$ ($\rho=0.2$ was excluded since it did not bring significant improvements in early trials). The node inference model $p_{\varepsilon}(n|\bm{y})$ was parameterized as a neural network with 2 layers with 512 units each and ReLU activations. 

\begin{table}[htb!]
\centering
\caption{Results on the QM9 dataset. First two columns are single conditioning, last column is multiple conditioning on $\mu$ and HOMO.
}
\label{tab:results_qm9}
\begin{tabular}{lcccccc}

\toprule
&\multicolumn{2}{@{}c@{}}{\textbf{$\mu$}}  &\multicolumn{2}{@{}c@{}}{\textbf{HOMO}} & \multicolumn{2}{@{}c@{}}{$\mu$+\textbf{HOMO}} \\

\cmidrule(lr){2-3} \cmidrule(lr){4-5} \cmidrule(lr){6-7}

& \textbf{MAE $\downarrow$} & \textbf{Val. $\uparrow$} & \textbf{MAE $\downarrow$} & \textbf{Val. $\uparrow$} & \textbf{MAE $\downarrow$} & \textbf{Val. $\uparrow$}\\
\midrule

\textbf{Unconditional} & $1.68 \pm 0.15$ & $91.5\%$ & $0.95 \pm 0.10$ & $91.5\%$ & $1.32 \pm 0.1$ & $91.5\%$\\
\midrule

\textbf{DiGress}\\

$\lambda = 200$  & $1.06 \pm 0.13$ & $87.0\%$ &$0.70 \pm 0.07$ & $91.7\%$ & $1.13 \pm 0.08$ & $88.7\%$\\
$\lambda = 400$  & $0.95 \pm 0.07$ & $86.5\%$ &$0.65 \pm 0.07$ & $91.7\%$ & $1.08 \pm 0.09$ & $87.2\%$\\
$\lambda = 600$  & $0.80 \pm 0.07$ & $82.5\%$ &$0.61 \pm 0.07$ & $91.2\%$ & $1.11 \pm 0.10$ & $86.7\%$\\
$\lambda = 800$  & $0.86 \pm 0.08$ & $76.7\%$ &$0.52 \pm 0.05$ & $90.5\%$ & $1.12 \pm 0.11$ & $87.1\%$\\
\midrule

\textbf{\textsc{FreeGress}}\\

$\rho = 0$ & $1.03 \pm 0.10$ & $86.7\%$ & $0.45 \pm 0.05$ & $92.0\%$ & $0.83 \pm 0.11$ & $82.3\%$\\
\cmidrule{1-1}

$\rho = 0.1, s = 1$ &$1.06 \pm 0.12$ & $88.3\%$ & $0.51 \pm 0.07$ & $92.3\%$ & $0.87 \pm 0.09$ & $86.2\%$\\
$\rho = 0.1, s = 3$ &$0.74 \pm 0.08$ & $83.7\%$ & $0.32 \pm 0.04$ & $90.1\%$ & $0.68 \pm 0.09$  & $83.4\%$\\
$\rho = 0.1, s = 5$ &$0.67 \pm 0.10$ & $77.7\%$ & $0.40 \pm 0.08$ & $92.9\%$ & $0.77 \pm 0.13$ & $77.0\%$\\
\cmidrule{1-1}

$\rho = 0.2, s = 1$ &$1.03 \pm 0.10$ & $89.1\%$ & $0.47 \pm 0.05$ & $91.6\%$ & $0.84 \pm 0.10$ & $84.3\%$\\
$\rho = 0.2, s = 3$ &$0.83 \pm 0.10$ & $82.1\%$ & $0.32 \pm 0.04$ & $93.8\%$ & $0.80 \pm 0.18$ & $72.0\%$\\
$\rho = 0.2, s = 5$ &$0.74 \pm 0.11$ & $75.5\%$ & $0.44 \pm 0.07$ & $93.3\%$ & $0.87 \pm 0.19$ & $72.7\%$\\

\midrule
\multicolumn{3}{l}{\textbf{\textsc{FreeGress}+extra features}}\\
$\rho = 0$           & $0.90 \pm 0.10$ & \bm{$90.1\%$} & $0.36 \pm 0.04$ & \bm{$94.2\%$} & $0.65 \pm 0.07$ & \bm{$93.1\%$}\\
\cmidrule{1-1}
$\rho = 0.1, s = 1$  & $0.90 \pm 0.10$ & $90.0\%$ & $0.35 \pm 0.03$ & $92.6\%$ & $0.65 \pm 0.07$ & $92.5\%$\\
$\rho = 0.1, s = 3$  & $0.60 \pm 0.07$ & $85.2\%$ & \bm{$0.27 \pm 0.04$} & $92.1\%$ & \bm{$0.54 \pm 0.07$} & $90.0\%$\\
$\rho = 0.1, s = 5$  & \bm{$0.52 \pm 0.07$} & $81.5\%$ & $0.40 \pm 0.08$ & $91.1\%$ & $0.59 \pm 0.11$ & $84.9\%$\\

\bottomrule

\end{tabular}
\end{table}

\section{Results}
\subsubsection{QM9.}
The experimental results on QM9 are presented in Table~\ref{tab:results_qm9}, where we report the MAE and the chemical validity of the generated molecules. Uniqueness scores are not reported as they approached 100\% in almost all cases, while the novelty is not a relevant metric on QM9 as the dataset is an enumeration of all the valid molecules satisfying a particular set of constraints. We observe that \textsc{FreeGress} variants, and especially those that leverage additional features, outperform DiGress variants in the vast majority of cases. Interestingly enough, even \textsc{FreeGress} variants with $\rho = 0$, i.e. those that use the guide without conditional dropout, are competitive with DiGress. Overall, \textsc{FreeGress} convincingly outperforms DiGress even when the additional features are not used (by 16\%, 38\% respectively on the $\mu$ and HOMO properties, and by 37\% when conditioning on both). When targeting $\mu$+HOMO in conjunction, the gap in performance indicates that \textsc{FreeGress} is better suited for multiple conditioning. As we show in Supplementary Table 1, this might be attributed to the property regressor of DiGress, which struggles to generate molecules conditioned on both properties, while \textsc{FreeGress} does not suffer from this issue. Overall, we observe a slight inverse correlation between MAE and validity, with one tending to decrease as the other improves. 
\begin{table}[htb!]
\centering
\caption{Results on the ZINC-250k dataset. First two columns are single conditioning, last column is multiple conditioning on LogP and QED.}
\label{tab:results_zinc_single}
\begin{tabular}{lcccccc}

\toprule
&\multicolumn{2}{@{}c@{}}{\textbf{LogP}}  &\multicolumn{2}{@{}c@{}}{\textbf{QED}} &\multicolumn{2}{@{}c@{}}{\textbf{LogP+QED}}\\
\cmidrule(lr){2-3} \cmidrule(lr){4-5} \cmidrule(lr){6-7}

& \textbf{MAE $\downarrow$} & \textbf{Val. $\uparrow$} & \textbf{MAE $\downarrow$} & \textbf{Val. $\uparrow$} & \textbf{MAE $\downarrow$} & \textbf{Val. $\uparrow$} \\
\midrule

\textbf{Unconditional} & $1.52 \pm 0.12$ & $86.1\%$ & $0.15 \pm 0.01$ & $86.1\%$ & $0.83 \pm 0.06$ & $86.1\%$\\
\midrule

\textbf{DiGress}\\
$\lambda = 200$  &$0.83 \pm 0.08$ & $77.6\%$ & $0.15 \pm 0.01$ & \bm{$85.8\%$} & $0.49 \pm 0.03$  & \bm{$83.8\%$}\\
$\lambda = 400$  &$0.74 \pm 0.08$ & $74.6\%$ & $0.15 \pm 0.01$ & $85.1\%$ & $0.40 \pm 0.02$ & $78.2\%$\\
$\lambda = 600$  &$0.76 \pm 0.10$ & $69.6\%$ & $0.14 \pm 0.01$ & $84.5\%$ & $0.38 \pm 0.02$ & $79.3\%$\\
$\lambda = 800$  &$0.92 \pm 0.12$ & $65.1\%$ & $0.14 \pm 0.01$ & $83.8\%$ & $0.35 \pm 0.01$ & $75.5\%$\\
\midrule

\textbf{\textsc{FreeGress}}\\
$\rho = 0$     & $0.22 \pm 0.02$ & $80.9\%$ & $0.06 \pm 0.01$ & $84.7\%$ & $0.15 \pm 0.01$ & $82.3\%$\\
\cmidrule{1-1}

$\rho = 0.1, s = 1$ &$0.22 \pm 0.02$ & $84.7\%$ & $0.06 \pm 0.01$ & $83.8\%$ & $0.16 \pm 0.01$ & $83.1\%$\\
$\rho = 0.1, s = 2$ &$0.17 \pm 0.01$ & $84.9\%$ & \bm{$0.04 \pm 0.01$} & $84.9\%$ & \bm{$0.12 \pm 0.01$} & $80.7\%$\\
$\rho = 0.1, s = 3$ &$0.18 \pm 0.02$ & $82.2\%$ & $0.05 \pm 0.01$ & $85.6\%$ & \bm{$0.12 \pm 0.01$} & $77.5\%$\\

\cmidrule{1-1}

$\rho = 0.2, s = 1$ & $0.22 \pm 0.02$ & \bm{$87.0\%$} & $0.07 \pm 0.01$ & $84.2\%$ & $0.17 \pm 0.02$ & $78.6\%$\\
$\rho = 0.2, s = 2$ & $0.19 \pm 0.02$ & $83.2\%$ & $0.05 \pm 0.01$ & $82.8\%$ & $0.13 \pm 0.01$ & $76.8\%$\\
$\rho = 0.2, s = 3$ & \bm{$0.16 \pm 0.01$} & $81.2\%$ & $0.05 \pm 0.01$ & $80.8\%$ & $0.13 \pm 0.02$ & $73.1\%$\\

\bottomrule
\end{tabular}
\end{table}

\subsubsection{ZINC-250k}
The experimental results of targeting logP and QED (alone and in conjunction) are presented in Table~\ref{tab:results_zinc_single}. Uniqueness and novelty scores are not reported as they approached 100\% in almost all cases. Similarly to QM9, we observe that \textsc{FreeGress} outperforms DiGress on all tasks by a wide margin. For example, on the LogP task, \textsc{FreeGress} with $\rho=0.2, s=3$ achieves a 78\% decrease in MAE with respect to the best DiGress variant ($\lambda=400$), while at the same time improving validity. On the QED task, \textsc{FreeGress} with $\rho=0.1, s=2$ achieves a 71\% MAE decrease with respect to the best DiGress variant ($\lambda=600$), while keeping the same validity rates. When conditioning on LogP and QED in conjunction, \textsc{FreeGress} with $\rho=0.1, s=2$ decreases MAE by 65\% in MAE compared to the best DiGress variant ($\lambda=800$) while slightly improving validity by 5\%. Overall, the results empirically confirm that \textsc{FreeGress} performs better than DiGress regardless of the dataset where it is applied and the properties that is tasked to target. 

\begin{table}[htb!]
\centering
\caption{Results of applying the proposed node inference method $p_{\varepsilon}(n\,|\,\bm{y})$ to \textsc{FreeGress} while targeting Molecular Weight on the ZINC-250k dataset. The last column reports the improvement against a ``no inference'' \textsc{FreeGress} variant. 
}
\label{tab:results_ni}
\begin{tabular}{lS[table-format=2.2 \pm 2.2]cS[table-format=2.2 \pm 2.2]ccc}

\toprule
&\multicolumn{2}{@{}c@{}}{\textbf{No inference}} &\multicolumn{2}{@{}c@{}}{$p_{\varepsilon}(n\,|\,\bm{y})$} &\multicolumn{2}{@{}c@{}}{\textbf{Improvement}}\\
\cmidrule(lr){2-3}\cmidrule(lr){4-5}\cmidrule(lr){6-7}

& \textbf{MAE $\downarrow$} & \textbf{Val. $\uparrow$} & \textbf{MAE $\downarrow$} & \textbf{Val. $\uparrow$} & \textbf{MAE $\downarrow$} & \textbf{Val. $\uparrow$}\\
\midrule

\textbf{DiGress}\\
$\lambda = 200$  & 67.01 \pm 6.20 & $\bm{82.5\%}$ & 19.75 \pm 2.27 & 38.7\% & -71.5\%  & -53.1\% \\
$\lambda = 400$  & 64.75 \pm 6.05 & 79.3\% & 22.51 \pm 2.62 & 38.4\% & -65.2\% & -51.6\% \\
$\lambda = 600$  & 64.67 \pm 6.17 & 80.2\% & 21.70 \pm 3.00 & 41.2\% & -66.4\% & -48.6\% \\
$\lambda = 800$  & 62.96 \pm 6.21 & 77.8\% & 20.92 \pm 2.90 & 40.4\% & -66.8\% & -48.1\% \\
\midrule

\textbf{\textsc{FreeGress}}\\
$\rho = 0$ & $\bm{7.08 \pm 1.19}$ & 54.5\% & 7.25 \pm 1.81 & 75.7\% & +2.4\% & +38.9\%\\
\cmidrule{1-1}

$\rho = 0.1, s = 1$ & 14.93 \pm 2.27 & 60.9\% & 8.22 \pm 0.77 & 81.4\% & -44.9\% & +33.7\%\\
$\rho = 0.1, s = 2$ & 11.37 \pm 2.81 & 50.8\% & 8.96 \pm 1.93 & 79.7\% & -21.2\% & +56.9\%\\
$\rho = 0.1, s = 3$ &  9.33 \pm 1.87 & 46.5\% & 8.03 \pm 1.02 & 66.6\% & -13.9\% & +43.2\%\\
\cmidrule{1-1}

$\rho = 0.2, s = 1$ & 20.00 \pm 5.47 & 61.3\% &  13.22 \pm 1.51 & 81.3\% & -33.9\% & +32.6\%\\
$\rho = 0.2, s = 2$ & 13.69 \pm 3.19 & 53.9\% &  13.37 \pm 1.47 & 78.4\% & -2.3\% & +45.5\%\\
$\rho = 0.2, s = 3$ & 15.61 \pm 4.08 & 48.4\% &  14.28 \pm 2.12 & 75.0\% & -8.5\% & +54.9\%\\

\cmidrule{1-7}
\multicolumn{3}{l}{\textbf{\textsc{FreeGress}+extra features}}\\
$\rho = 0$          & 12.66 \pm 4.99 & 60.1\% & 5.13 \pm 0.38 & $\bm{86.1\%}$ & -91.85\% & +43.25\%\\
\cmidrule{1-1}
$\rho = 0.1, s = 1$ & 43.27 \pm 9.35 & 82.1\% &  3.52 \pm 0.34 & 83.7\%  & -91.87\% & +1.95\%\\
$\rho = 0.1, s = 2$ & 48.25 \pm 11.03 & 73.3\% &  $\bm{3.34 \pm 0.31}$   &   84.7\% & -93.08\% & +15.55\%\\
$\rho = 0.1, s = 3$ & 47.67 \pm 13.26 & 63.2\% &  3.35 \pm 0.40 & 81.2\% & -92.97\% & +28.48\%\\

\bottomrule
\end{tabular}
\end{table}

\subsubsection{Effect of node inference in ZINC-250k}
Table~\ref{tab:results_ni} summarizes the results on applying node inference on the MW targeting task. The ``no inference'' scenario (first column) indicates that \textsc{FreeGress} obtains lower MAEs than DiGress, regardless of whether additional features are added or not. However, this happens at the expense of validity, which is lower for \textsc{FreeGress} than DiGress. Adding node inference (central column) allows \textsc{FreeGress} to generate valid molecules at the same rate than DiGress, but with better MAE. Moreover, the addition of node inference is beneficial to \textsc{FreeGress} itself, as quantified by the third column which measures row-wise improvement. Surprisingly, we observe that applying node inference to DiGress decreases MAE by approximately 71\% across experiments, but at the same time completely disrupts chemical validity. In contrast, adding node inference to \textsc{FreeGress} causes a good decrease in MAE and a steep increase in validity (up to 57\%). Overall, the best improvement is given by the combination of node inference and additional molecular features. 

\begin{figure}[ht]
	\minipage[t]{0.24\columnwidth}
        \centering
        \captionsetup{justification=centering}
        \includegraphics[width=2cm]{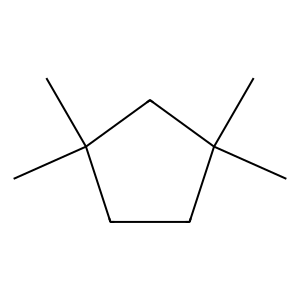}
        \caption*{\scriptsize{Input $\mu$: 0.0603\\Est. $\mu$: 0.0463}}
	\endminipage
	\minipage[t]{0.24\columnwidth}
        \centering
        \captionsetup{justification=centering}
        \includegraphics[width=2cm]{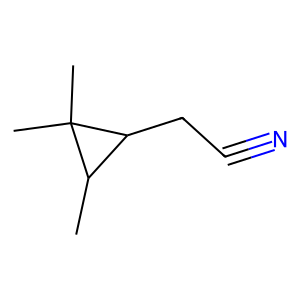}
        \caption*{\scriptsize{Input $\mu$: 4.2338\\Est. $\mu$: 4.1238}}
	\endminipage
	\minipage[t]{0.24\columnwidth}
        \centering
        \captionsetup{justification=centering}
        \includegraphics[width=2cm]{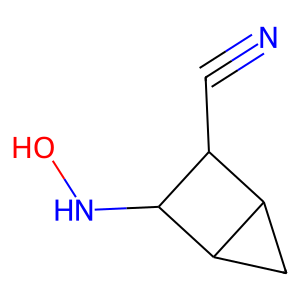}
        \caption*{\scriptsize{Input HOMO: -6.8083\\Est. HOMO: -6.8604}}
	\endminipage
	\minipage[t]{0.24\columnwidth}
        \centering
        \captionsetup{justification=centering}
        \includegraphics[width=2cm]{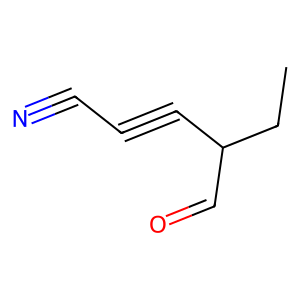}
        \caption*{\scriptsize{Input HOMO: -7.4559\\Est. HOMO: -7.4632}}
	\endminipage
        \\
	\minipage[t]{0.24\columnwidth}
        \centering
        \captionsetup{justification=centering}
        \includegraphics[width=2.5cm]{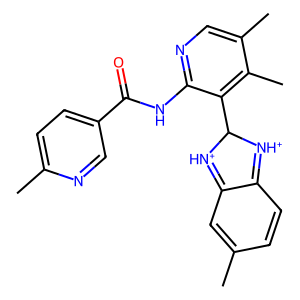}
        \caption*{\scriptsize{Input LogP: -0.9834\\Est. LogP: -0.9246}}
	\endminipage
	\minipage[t]{0.24\columnwidth}
        \centering
        \captionsetup{justification=centering}
        \includegraphics[width=2.5cm]{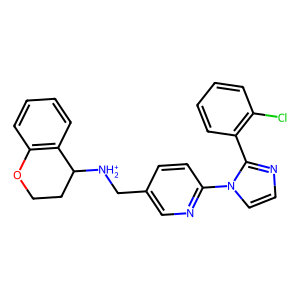}
        \caption*{\scriptsize{Input QED: 0.5034\\Est. QED: 0.5297}}
	\endminipage
	\minipage[t]{0.24\columnwidth}
        \centering
        \captionsetup{justification=centering}
        \includegraphics[width=2.5cm]{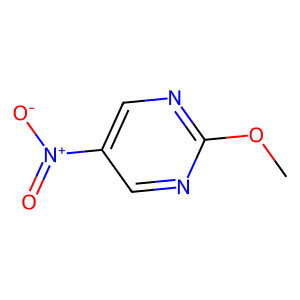}
        \caption*{\scriptsize{Input MW: 159.05\\Est. MW: 155.03}}
	\endminipage
	\minipage[t]{0.24\columnwidth}
        \centering
        \captionsetup{justification=centering}
        \includegraphics[width=2.5cm]{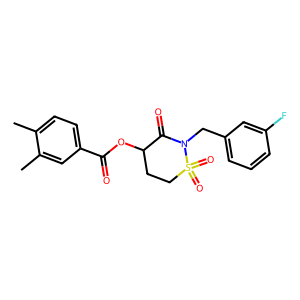}
        \caption*{\scriptsize{Input MW: 401.13\\Est. MW: 405.10}}
	\endminipage
    \caption{Curated molecules from the QM9 (top row) and ZINC-250k (bottom row). The input conditioning value and the one estimated (Est.) after generation are displayed below each molecule.}\label{fig:samples}
\end{figure}

\subsubsection{Curated samples.} 
In Figure~\ref{fig:samples} we show some curated samples from the targeting properties under study under study to add visual evidence that the model is operating correctly and it is able to target different properties with success. Additional random samples are shown in the Supplementary material.

\section{Conclusions}
We introduced \textsc{FreeGress}, a discrete DDPM for graphs with CF guidance, which allows to generate molecules complying with a pre-specified set of desired chemical properties. Through extensive experiments, we have shown that CF guidance allows to generate better (more tailored to the specification) molecules than CB guidance without sacrificing chemical validity. Additionally, we have implemented a form of learned node inference, and shown that inferring the number of nodes from the guide helps the generation whenever the molecular size is connected to the target property.

In future studies, we intend to tackle some of the current limitations of \textsc{FreeGress} to expand its applicability. A first research direction is to improve chemical validity designing the forward and reverse processes to use conditional transition matrices, thus constraining the intermediate noising and denoising steps to happen in valid chemical space. Another promising direction is to work with molecular fragments rather than atoms, as it would likely reduce the opportunities to generate an invalid intermediate structure. We intend to validate these findings also in the more general setting of arbitrary graph generation. 

\begin{credits}
\subsubsection{\ackname} Research partly funded by PNRR - M4C2 - Investimento 1.3, Partenariato Esteso PE00000013 - "FAIR - Future Artificial Intelligence Research" - Spoke 1 "Human-centered AI", funded by the European Commission under the NextGeneration EU programme.

\subsubsection{\discintname}
The authors have no competing interests to declare that are relevant to the content of this article.
\end{credits}
\bibliographystyle{splncs04}
\bibliography{original_references}

\end{document}

\begin{center}
    \begin{Large}\textbf{SUPPLEMENTARY MATERIAL}\end{Large}\\
    \vspace{.5cm}
    \begin{Large}\textbf{Classifier-free graph diffusion\\for molecular property targeting}\end{Large}
\end{center}
Here, we provide some data on the datasets employed, as well as some additional information on the experiments executed.

\section{Datasets}
Below is some additional information on the QM9 and ZINC-250k datasets, such as training, validation, and test sizes, as well as some additional dataset statistics.
\paragraph{QM9}
\begin{itemize}
    \item Training set size: 100000 
    \item Validation set size: 20497 
    \item Test set size: 100 (the actual size is 13388)
    \item Most frequent number of nodes/atoms: 9
    \item Average number of atoms: 8.79
    \item Average number of bonds: 9.40
\end{itemize}
\paragraph{ZINC-250k}
\begin{itemize}
    \item Training set size: 199564 (out of which 183475 passed the preprocessing step)
    \item Validation set size: 24946 (out of which 22788 passed the preprocessing step)
    \item Test set size: 100 (the actual size is 22629)
    \item Most frequent number of nodes/atoms: 24
    \item Average number of atoms: 23.15
    \item Average number of bonds: 24.90
\end{itemize}

\pagebreak
\section{Additional results}
The following tables provide additional insights into the experiments performed on the main document where we conditioned on two properties at the same time. We provide the individual Mean Absolute Error for each of the properties considered. As it is possible to see, it appears that DiGress struggles to work with two properties at the same time, while FreeGress does not seem to have this kind of issue.

\begin{table}[htb!]
\centering
\caption{In-depth results over $\mu$ and HOMO.}
\begin{tabular}{lcccc}

\toprule
&\multicolumn{4}{@{}c@{}}{\textbf{$\mu$+HOMO}}\\

\midrule

& \textbf{$\mu$ MAE $\downarrow$} & \textbf{HOMO MAE $\downarrow$} & \textbf{Total MAE $\downarrow$} & \textbf{Val. $\uparrow$}\\
\midrule

\textbf{Unconditional} & $1.68 \pm 0.16$ & $0.96 \pm 0.1$ & $1.32 \pm 0.1$ & $91.5\%$\\
\midrule

\textbf{DiGress}\\
%Original results & $0.81 \pm 0.04$ & N.A. &$0.56 \pm 0.01$ & N.A. & N.A. & N.A. & $0.87$ & N.A. \\

$\lambda = 200$  & $1.25 \pm 0.10$ & $1.01 \pm 0.10$ & $1.13 \pm 0.08$ & $88.7\%$\\
$\lambda = 400$  & $1.10 \pm 0.13$ & $1.05 \pm 0.12$ & $1.08 \pm 0.09$ & $87.2\%$\\
$\lambda = 600$  & $1.09 \pm 0.14$ & $1.13 \pm 0.13$ & $1.11 \pm 0.10$ & $86.7\%$\\
$\lambda = 800$  & $1.08 \pm 0.13$ & $1.16 \pm 0.14$ & $1.12 \pm 0.11$ & $87.1\%$\\
\midrule

\textbf{\textsc{FreeGress}}\\

$\rho = 0$ & $0.95 \pm 0.11$ & $0.35 \pm 0.04$ & $0.83 \pm 0.11$ & $82.3\%$\\
\cmidrule{1-1}

$\rho = 0.1, s = 1$ & $1.08 \pm 0.11$ & $0.66 \pm 0.12$ & $0.87 \pm 0.09$ & $86.2\%$\\
$\rho = 0.1, s = 3$ & $0.85 \pm 0.11$ & $0.52 \pm 0.12$ & $0.68 \pm 0.09$ & $83.4\%$\\
$\rho = 0.1, s = 5$ & $0.88 \pm 0.13$ & $0.67 \pm 0.19$ & $0.77 \pm 0.13$ & $77.0\%$\\
\cmidrule{1-1}

$\rho = 0.2, s = 1$ & $1.06 \pm 0.12$ & $0.62 \pm 0.11$ & $0.84 \pm 0.10$ & $84.3\%$\\
$\rho = 0.2, s = 3$ & $0.92 \pm 0.13$ & $0.68 \pm 0.18$ & $0.80 \pm 0.18$ & $72.0\%$\\
$\rho = 0.2, s = 5$ & $0.95 \pm 0.17$ & $0.79 \pm 0.23$ & $0.87 \pm 0.19$ & $72.7\%$\\

\midrule
\multicolumn{3}{l}{\textbf{\textsc{FreeGress}+extra features}}\\
$\rho = 0$           & $0.95 \pm 0.11$ & $0.35 \pm 0.04$ & $0.65 \pm 0.07$  & \bm{$93.1\%$}\\
\cmidrule{1-1}
$\rho = 0.1, s = 1$  & $0.96 \pm 0.12$ & \bm{$0.34 \pm 0.04$} & $0.65 \pm 0.07$  & $92.5\%$\\
$\rho = 0.1, s = 3$  & \bm{$0.74 \pm 0.10$} & \bm{$0.34 \pm 0.06$} & \bm{$0.54 \pm 0.07$} & $90.0\%$\\
$\rho = 0.1, s = 5$  & $0.75 \pm 0.10$ & $0.42 \pm 0.15$ & $0.59 \pm 0.11$  & $84.9\%$\\

\bottomrule
\end{tabular}
\end{table}

\begin{table}[!htp]\centering
\centering
\caption{In-depth results over multiple conditioning on LogP and QED.}
\begin{tabular}{lcccc}

\toprule
&\multicolumn{4}{@{}c@{}}{\textbf{LogP+QED}}\\
\midrule

& \textbf{LogP MAE $\downarrow$} & \textbf{QED MAE $\downarrow$} & \textbf{Total MAE $\downarrow$} & \textbf{Val. $\uparrow$} \\
\midrule

\textbf{Unconditional} & $1.51 \pm 0.12$ & $0.15 \pm 0.01$ & $0.83 \pm 0.06$ & $86.1\%$\\
\midrule

\textbf{DiGress}\\
$\lambda = 200$  & $0.83 \pm 0.05$ & $0.14 \pm 0.01$ & $0.49 \pm 0.03$  & \bm{$83.8\%$}\\
$\lambda = 400$  & $0.67 \pm 0.04$ & $0.13 \pm 0.01$ & $0.40 \pm 0.02$ & $78.2\%$\\
$\lambda = 600$  & $0.63 \pm 0.04$ & $0.14 \pm 0.01$ & $0.38 \pm 0.02$ & $79.3\%$\\
$\lambda = 800$  & $0.55 \pm 0.03$ & $0.14 \pm 0.01$ & $0.35 \pm 0.01$ & $75.5\%$\\
\midrule

\textbf{\textsc{FreeGress}}\\
$\rho = 0$ & $0.23 \pm 0.02$  & $0.07 \pm 0.01$  & $0.15 \pm 0.01$ & $82.3\%$\\
\cmidrule{1-1}

$\rho = 0.1, s = 1$ & $0.24 \pm 0.02$ & $0.07 \pm 0.01$ & $0.16 \pm 0.01$ & $83.1\%$\\
$\rho = 0.1, s = 2$ & $0.19 \pm 0.02$ & $0.06 \pm 0.01$ & \bm{$0.12 \pm 0.01$} & $80.7\%$\\
$\rho = 0.1, s = 3$ & \bm{$0.18 \pm 0.02$} & \bm{$0.05 \pm 0.01$} & \bm{$0.12 \pm 0.01$} & $77.5\%$\\
\cmidrule{1-1}

$\rho = 0.2, s = 1$ & $0.26 \pm 0.02$ & $0.06 \pm 0.01$ & $0.17 \pm 0.02$ & $78.6\%$\\
$\rho = 0.2, s = 2$ & $0.20 \pm 0.02$ & $0.06 \pm 0.06$ & $0.13 \pm 0.01$ & $76.8\%$\\
$\rho = 0.2, s = 3$ & $0.20 \pm 0.02$ & $0.07 \pm 0.01$ & $0.13 \pm 0.02$ & $73.1\%$\\

\bottomrule
\end{tabular}
\end{table}

\pagebreak

\section{Additional molecules}
We provide a list of ten randomly generated molecules for each dataset considered in this study. Below each molecule, the targeted property (Input) and the property estimated on the generated molecule (Est.) are shown.
\begin{figure}[ht]
    % \captionsetup{justification=centering}
	\minipage[t]{0.2\columnwidth}
        \centering
        \captionsetup{justification=centering}
        \includegraphics[width=2cm]{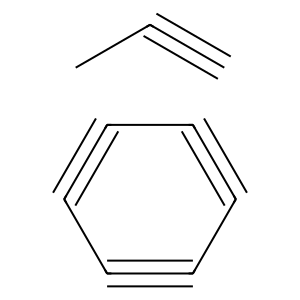}
        \caption*{\scriptsize{Input $\mu$: 0\\Est. $\mu$: 1.8289}}
	\endminipage
	\minipage[t]{0.2\columnwidth}
        \centering
        \captionsetup{justification=centering}
        \includegraphics[width=2cm]{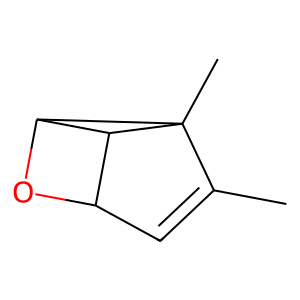}
        \caption*{\scriptsize{Input $\mu$: 1.7341\\Est. $\mu$: 2.3256}}
	\endminipage
    \minipage[t]{0.2\columnwidth}
        \centering
        \captionsetup{justification=centering}
        \includegraphics[width=2cm]{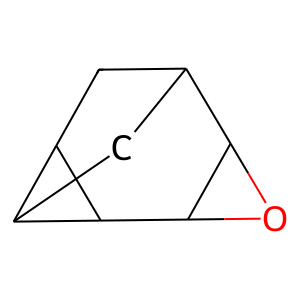}
        \caption*{\scriptsize{Input $\mu$: 1.8511\\Est. $\mu$: 2.0197}}
	\endminipage
    \minipage[t]{0.2\columnwidth}
        \centering
        \captionsetup{justification=centering}
        \includegraphics[width=2cm]{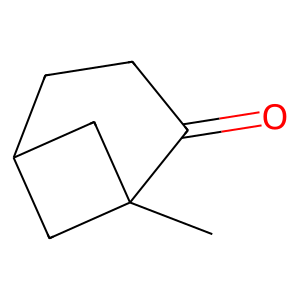}
        \caption*{\scriptsize{Input $\mu$: 2.8579\\Est. $\mu$: 2.9735}}
	\endminipage
    \minipage[t]{0.2\columnwidth}
        \centering
        \captionsetup{justification=centering}
        \includegraphics[width=2cm]{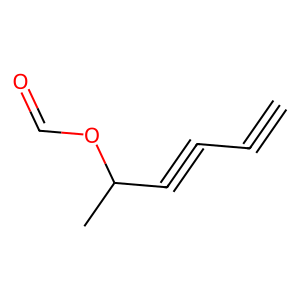}
        \caption*{\scriptsize{Input $\mu$: 3.9200\\Est. $\mu$: 1.9186}}
	\endminipage
    \\
	\minipage[t]{0.2\columnwidth}
        \centering
        \captionsetup{justification=centering}
        \includegraphics[width=2cm]{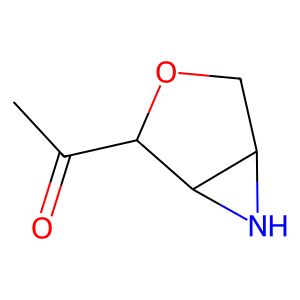}
        \caption*{\scriptsize{Input HOMO: -6.5634\\Est. HOMO: -6.2288}}
	\endminipage
	\minipage[t]{0.2\columnwidth}
        \centering
        \captionsetup{justification=centering}
        \includegraphics[width=2cm]{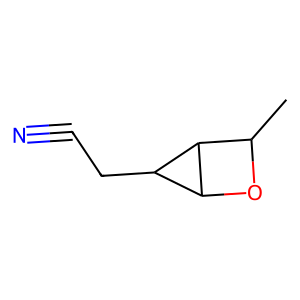}
        \caption*{\scriptsize{Input HOMO: -6.9035\\Est. HOMO: -6.7593}}
	\endminipage
    \minipage[t]{0.2\columnwidth}
        \centering
        \captionsetup{justification=centering}
        \includegraphics[width=2cm]{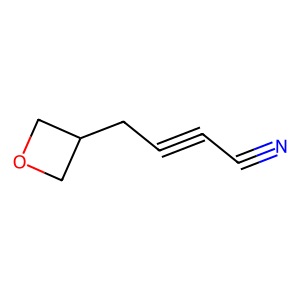}
        \caption*{\scriptsize{Input HOMO: -7.2219\\Est. HOMO: -7.0610}}
	\endminipage
    \minipage[t]{0.2\columnwidth}
        \centering
        \captionsetup{justification=centering}
        \includegraphics[width=2cm]{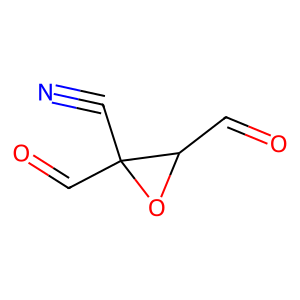}
        \caption*{\scriptsize{Input HOMO: -7.6573\\Est. HOMO: -7.8576}}
	\endminipage
    \minipage[t]{0.2\columnwidth}
        \centering
        \captionsetup{justification=centering}
        \includegraphics[width=2cm]{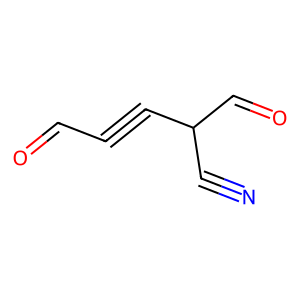}
        \caption*{\scriptsize{Input HOMO: -7.9675\\Est. HOMO: -7.8117}}
	\endminipage
    \caption{Non-curated molecules from the QM9 dataset.}
\end{figure}

\begin{figure}[ht]
    % \captionsetup{justification=centering}
	\minipage[t]{0.2\columnwidth}
        \centering
        \captionsetup{justification=centering}
        \includegraphics[width=2cm]{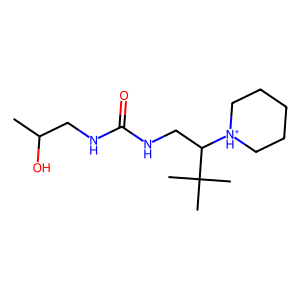}
        \caption*{\scriptsize{Input LogP: 0.0964\\Est. LogP: 0.1499}}
	\endminipage
	\minipage[t]{0.2\columnwidth}
        \centering
        \captionsetup{justification=centering}
        \includegraphics[width=2cm]{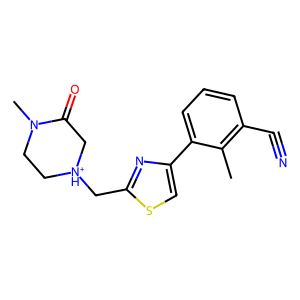}
        \caption*{\scriptsize{Input LogP: 0.8472\\Est. LogP: 0.8471}}
	\endminipage
    \minipage[t]{0.2\columnwidth}
        \centering
        \captionsetup{justification=centering}
        \includegraphics[width=2cm]{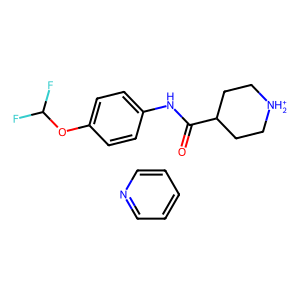}
        \caption*{\scriptsize{Input LogP: 2.1313\\Est. LogP: 2.2815}}
	\endminipage
    \minipage[t]{0.2\columnwidth}
        \centering
        \captionsetup{justification=centering}
        \includegraphics[width=2cm]{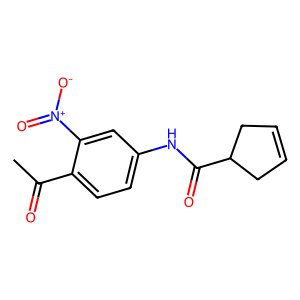}
        \caption*{\scriptsize{Input QED: 0.5034\\Est. QED: 0.3954}}
	\endminipage
    \minipage[t]{0.2\columnwidth}
        \centering
        \captionsetup{justification=centering}
        \includegraphics[width=2cm]{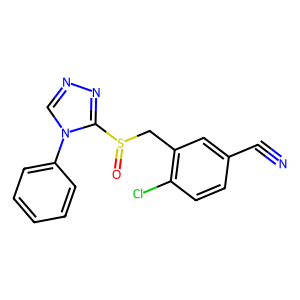}
        \caption*{\scriptsize{Input QED: 0.6735\\Est. QED: 0.7300}}
	\endminipage
    \\
	\minipage[t]{0.2\columnwidth}
        \centering
        \captionsetup{justification=centering}
        \includegraphics[width=2cm]{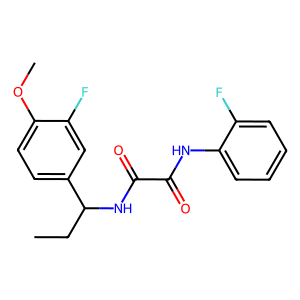}
        \caption*{\scriptsize{Input QED: 0.7246\\Est. QED: 0.8158}}
	\endminipage
	\minipage[t]{0.2\columnwidth}
        \centering
        \captionsetup{justification=centering}
        \includegraphics[width=2cm]{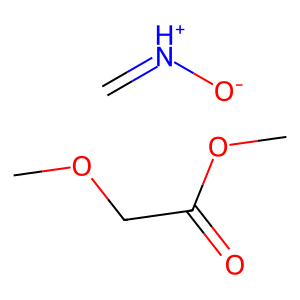}
        \caption*{\scriptsize{Input MW: 159.05\\Est. MW: 149.07}}
	\endminipage
    \minipage[t]{0.2\columnwidth}
        \centering
        \captionsetup{justification=centering}
        \includegraphics[width=2cm]{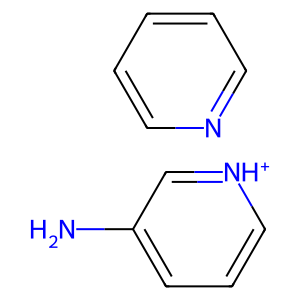}
        \caption*{\scriptsize{Input MW: 175.12\\Est. MW: 174.10}}
	\endminipage
    \minipage[t]{0.2\columnwidth}
        \centering
        \captionsetup{justification=centering}
        \includegraphics[width=2cm]{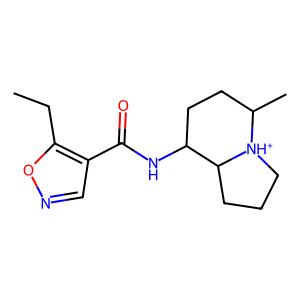}
        \caption*{\scriptsize{Input MW: 279.09\\Est. MW: 278.19}}
	\endminipage
    \minipage[t]{0.2\columnwidth}
        \centering
        \captionsetup{justification=centering}
        \includegraphics[width=2cm]{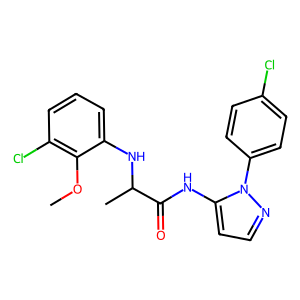}
        \caption*{\scriptsize{Input MW: 400.04\\Est. MW: 404.08}}
	\endminipage
    \caption{Non-curated molecules from the ZINC-250k dataset.}
\end{figure}